\documentclass[letterpaper, 10 pt, conference]{ieeeconf}

\IEEEoverridecommandlockouts                              

\usepackage{graphics} 
\usepackage{epsfig} 
\usepackage{amsmath} 
\usepackage{amssymb}  
\usepackage{multirow}
\usepackage{xcolor}
\usepackage{caption}
\usepackage{graphicx}
\usepackage{subcaption}
\usepackage{mathtools}
\usepackage{multirow}
\usepackage[ruled,vlined,linesnumbered]{algorithm2e}
\usepackage{tabularx}
\usepackage{soul}
\usepackage[font=small]{caption}

\newlength\mylen

\DeclareMathOperator*{\argmax}{arg\,max}

\newcommand{\reward}[1]{(^{#1} r)^i}

\title{\LARGE \bf Learning to Coordinate for a \textit{Worker-Station} Multi-robot System \\ in Planar Coverage Tasks}

\author{
Jingtao Tang$^{1}$,
Yuan~Gao$^{2}$,
Tin Lun Lam$^{1,2\dagger} $
\thanks{$^{1}$Authors are with School of Science and Engineering, The Chinese University of Hong Kong, Shenzhen. Authors' email is todd.j.tang@gmail.com.}
\thanks{$^{2}$Authors are with Shenzhen Institute of Artificial Intelligence and Robotics for Society, Shenzhen, Guang Dong, China. Authors' emails are gaoyuankidult@gmail.com and tllam@cuhk.edu.cn.}
\thanks{We gratefully acknowledge support of the National Key R\&D Program of China (2020YFB1313300) and the funding (AC01202101025,
AC01202101026) from the Shenzhen Institute of Artificial Intelligence
and Robotics for Society.}
\thanks{$^\dagger$Corresponding author}
}

\begin{document}

\maketitle
\thispagestyle{empty}
\pagestyle{empty}

\begin{abstract}
For massive large-scale tasks, a multi-robot system (MRS) can effectively improve efficiency by utilizing each robot's different capabilities, mobility, and functionality.
In this paper, we focus on the multi-robot coverage path planning (mCPP) problem in large-scale planar areas with random dynamic interferers in the environment, where the robots have limited resources.
We introduce a \textit{worker-station} MRS consisting of multiple \textit{workers} with limited resources for actual work, and one \textit{station} with enough resources for resource replenishment.
We aim to solve the mCPP problem for the \textit{worker-station} MRS by formulating it as a fully cooperative multi-agent reinforcement learning problem.
Then we propose an end-to-end decentralized online planning method, which simultaneously solves coverage planning for \textit{workers} and rendezvous planning for \textit{station}.
Our method manages to reduce the influence of random dynamic interferers on planning, while the robots can avoid collisions with them.
We conduct simulation and real robot experiments, and the comparison results show that our method has competitive performance in solving the mCPP problem for \textit{worker-station} MRS in metric of task finish time.
\end{abstract} 
\section{Introduction} \label{sec:intro}

For massive large-scale tasks in hazardous environments, Multi-Robot System (MRS) greatly helps to reduce human exposure to potential dangers and improves efficiency effectively.
There are various applications of MRS that have come to reality, including search and rescue~\cite{liu2016multirobot}, persistent surveillance~\cite{palacios2016distributed}, planetary exploration~\cite{schuster2020arches}.
Typically, a robot has only limited working resources, including energy and consumables.
For example, most robots are driven by electrical or thermal energy stored in batteries or fuels in advance.
While some robots can obtain ambient energy from the environment (e.g., solar energy), the energy transfer is highly dependent on the environmental situations, and the recharging process can sometimes be slow. Thus it can be inefficient for robots in massive long-term tasks.
In scenarios like cleaning or agriculture on large-scale fields, the robot has limited consumables like water or chemicals.
Therefore, it is essential for robots with large-scale tasks to constantly travel between supply stations and working areas to replenish and work, which is very inefficient for such tasks.
\begin{figure}[t]
\centering
\includegraphics[width=0.9\linewidth]{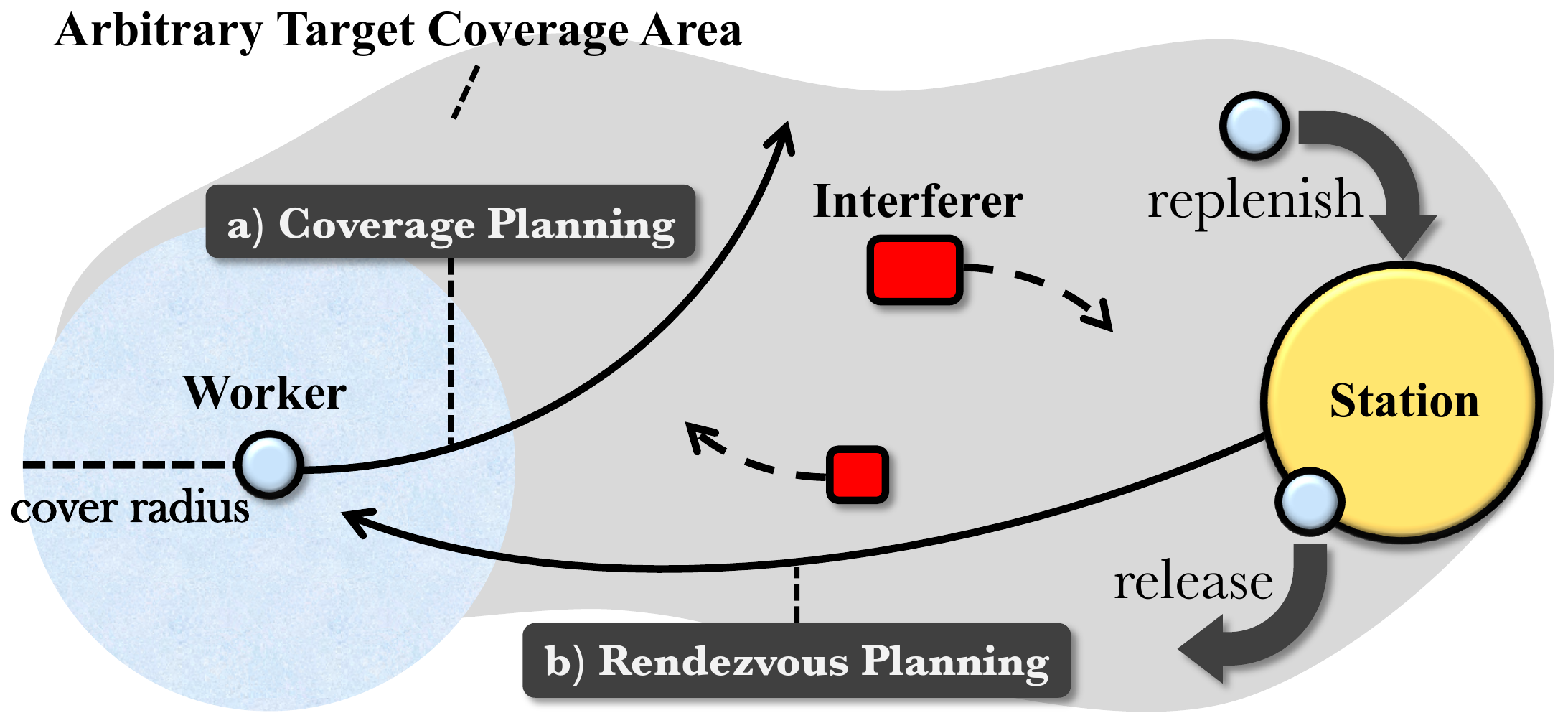}
\caption{Given an arbitrary target coverage area, a mCPP problem for the \textit{worker-station} MRS on planar areas can be decomposed into: 1) coverage planning for \textit{workers} (blue robots) and 2) rendezvous planning for \textit{station} (yellow robot). There are random dynamic interferers (red robots) in the environment.}
\label{fig:problem_demo}
\vspace{-5mm}
\end{figure}

As discussed by Vaughan et al. in \cite{couture2009adaptive}, the placement of the supply station significantly influences work efficiency for an MRS in the above scenarios.
To further improve the efficiency of the MRS, one might consider making the supply station a mobile robot platform.
Similar to the \textit{Frugal Feeding Problem} in \cite{litus2007frugal}, the station moves around to serve the working robots.
For consistency in this paper, we name such an MRS as the "\textit{worker-station}" MRS, which is composed of a mobile supply station robot and several working robots.
We consider the Multi-robot Coverage Path Planning (mCPP) problem on planar areas for the aforementioned \textit{worker-station} MRS.
As shown in Fig.~\ref{fig:problem_demo}, the \textit{workers} are equipped with a range device for general area coverage work, and the \textit{station} is loaded with sufficient resources to provide supplies for \textit{workers}.
The joint objective is to cover a given target area as soon as possible.
Solving such a planning problem for the \textit{worker-station} MRS can be decomposed as below:
\begin{enumerate}
    \item Coverage planning for each \textit{worker} to finish general planar coverage work of a given area;
    \item Rendezvous planning for the \textit{station} to service \textit{workers} in need of replenishment;
\end{enumerate}

In this paper, we mainly focus on solving the mCPP problem for the \textit{worker-station} MRS on planar areas.
There are several challenges to the above problem.
First, the joint problem space comprised of the above two planning problems is too large to solve directly and simultaneously.
A practical solution is to discretize state and action spaces in mCPP problems \cite{almadhoun2019survey} and rendezvous planning problems \cite{litus2008distributed}, then solve by discrete combinatorial optimization methods separately.
However, the system dynamics are hard to model and identify, where each robot has different capabilities and functionality.
Thus, such methods can still be infeasible for such complex MRS, even after reducing the problem size.
Secondly, planning with dynamic collision avoidance is another challenge for most offline planning methods.
One general solution is to combine offline planning with local collision avoidance controllers.
Nevertheless, such a hierarchical planning scheme would alter the optimal policy that is planned offline without the interference of dynamic obstacles.
For complex scheduling tasks like the mCPP problem for \textit{worker-station} MRS, it will cause conflicts and even deadlocks between robots or planners, and the planning efficiency will get worse as the number of robots grows~\cite{wang2017safety}.
To tackle the above challenges, we adopt Deep Reinforcement Learning (DRL) to solve the mCPP problem for \textit{worker-station} MRS.
However, the coordination behaviors of different agents in the \textit{worker-station} MRS are nontrivial to learn together, and agents often struggle between exploration and exploitation of the coverage task during training.
We summarize the main contributions of this paper below:
\begin{enumerate}
\item
We propose an end-to-end decentralized online planning method to solve the mCPP problem for the \textit{worker-station} MRS.
Our method manages to reduce the influence of random dynamic interferers on planning, while the robots can avoid collisions with them.
\item
We design a two-stage curriculum learning with an intrinsic curiosity module and soft approximation of the \textit{workers}' energy constraints, which successfully guide the training for large-scale coverage tasks.
\item
We provide ablation study, simulation, and real robot experimental results.
The results show that our method outperforms decomposition-based and graph-based baseline methods in coverage finish time metrics.
\end{enumerate}

\section{Related Work} \label{sec:related-work}
\subsection{Multi-robot Coverage Path Planning}
The mCPP problem evolved from the classical Coverage Path Planning (CPP) problem by introducing multiple robots to solve the coverage problem.
Most approaches are based on the graph structure, which is proven to be NP-hard \cite{MFC-IROS'05}.
Zheng \textit{et al.} designed a constant-factor approximation algorithm in polynomial time \cite{MFC-2010}.
Kapoutsis \textit{et al.} uses an area division algorithm to allocate tasks for multiple robots \cite{kapoutsis2017darp}.
Apart from graph-based methods, decomposition-based methods also take large parts in the literature~\cite{rekleitis2008efficient, collins2021scalable}, which first partition the target area into obstacle-free convex sub-regions for different robots and then apply single robot coverage planning for each robot separately.
Most graph-based or decomposition-based mCPP methods do offline planning, and some also require the coverage area to satisfy specific assumptions (e.g., convex-shaped area).
In addition, classic offline mCPP methods can be undermined by random dynamic interferers in the environment.

On the other hand, recently, some works have been extending the mCPP problem to various applications with specific constraints, such as geophysical surveys~\cite{azpurua2018multi}, fault-tolerant planning on large-scale outdoor~\cite{sun2021ft}.
However, to the best of our knowledge, there are few works on the mCPP problem for the aforementioned \textit{worker-station} MRS.

\subsection{\textit{Worker-Station} Multi-robot System}
Similar to the \textit{worker-station} MRS, related works on mobilizing the supply station into an autonomous robot mainly focus on rendezvous planning for \textit{station} to efficiently recharge the \textit{workers} in need.
For example, Couture et al. \cite{couture2009adaptive} only plans for \textit{station}, while \textit{workers} are dedicated to delivering goods between fixed source and destination.
Similarly, in \cite{mathew2015multirobot}, only rendezvous planning of \textit{stations} is considered, whereas the \textit{workers} are programmed to monitor the environment by predefined trajectories persistently.
Most of these works consider the \textit{workers} to be stationary in terms of their motion patterns and state transitions, which reduce the complexity to a solvable level for optimization.

More recently, Yu et al. \cite{yu2018algorithms} tried to solve both planning problems for one \textit{worker} and one \textit{station}, but it is restricted to node coverage for a given graph.
Similarly in Sun et al. \cite{sun2020unified}, the \textit{worker} is planned to travel between waypoints, while the \textit{station} is planned to rendezvous to charge the \textit{worker}.
Seyedi et al. \cite{seyedi2019persistent} planned trajectories for multiple \textit{workers} and one \textit{station} with scheduled charging order, which also requires a prior of the environment.
Despite the above work managed to plan for both \textit{workers} and \textit{station}, it is only applicable on convex target areas with static obstacles, thus is infeasible in an arbitrary target area with dynamic interferers.

\newcommand{\action}{a}
\newcommand{\position}{\boldsymbol{x}}
\newcommand{\robot}[1]{R^{#1}}
\newcommand{\worker}[1]{W^{#1}}
\newcommand{\station}[1]{S^{#1}}
\newcommand{\capacity}[1]{c^{#1}}
\newcommand{\coverage}{\mathcal{C}}

\section{Problem Formulation}\label{sec:problem}
In this section, we provide our Multi-agent Reinforcement Learning (MARL) problem formulation of the Multi-robot Coverage Path Planning (mCPP) problem on planar areas, for the \textit{worker-station} Multi-robot System (MRS) (see Fig.~\ref{fig:problem_demo}).
Given target area $\Omega$, the \textit{worker-station} MRS consists of $m$ \textit{workers} $\worker{i}$ and $n$ \textit{stations} $\station{j}$, where $i=1,2,...,m$ and $j=1,2,...,n$.
The goal is to find the optimal policy for each robot in the \textit{worker-station} MRS, to minimize the coverage task finish time while avoiding collisions with dynamic interferers in the environment.

\subsection{Preliminaries}
In this subsection, we first introduce several preliminary concepts and assumptions in the rest of the paper.

\subsubsection{\textit{Worker} robot and \textit{station} robot}
we consider both \textit{workers} and \textit{station} have limited range of perception and communication:
within the perception range, each robot can detect collisions and objects precisely; within the communication range, each robot can receive information from other robots (e.g., the rough global position of other robots).
As mentioned previously in the \textit{worker-station} MRS, \textit{workers} also have limited energy, while \textit{station} have unlimited energy to replenish \textit{workers}.
Note that the coverage work range of \textit{worker} does not necessarily equal its perception range.  

\subsubsection{Energy Capacity and Rendezvous Recharge}
denote the energy capacity for \textit{worker} $\worker{i}$ as a constant $\capacity{i}$.
Suppose the current energy left for $\worker{i}$ at time $t$ is $e_t^i$, then current percentage of remained energy $p_t^i$ is defined as: $p_t^i=\frac{e_t^i}{\capacity{i}}\in[0, 1]$.
A \textit{worker} $\worker{i}$ is said to be ``\textit{exhausted}'' if $p_t^i$ is lower than a threshold $p^e$, otherwise it is said to be ``\textit{normal}''.
Also, since we mainly focus on the planning problem at a higher level in this paper, the local rendezvous of \textit{workers} and \textit{station} is simplified by comparing with a position threshold $\varepsilon$.
We assume a \textit{worker} can be replenished by any \textit{station}, then the discharge and recharge for each \textit{worker} $\worker{i}$ is determined by comparing $\varepsilon$ with the euclidean distance between the global positions of $\position_t^{\worker{i}}$ and $\position_t^{\station{j}}$ for \textit{worker} $\worker{i}$ and \textit{station} $\station{j}$:
\begin{equation}\label{eq:energy_model}
e_t^i = \begin{cases} \text{max}\{0,\;e_{t-1}^i - e_{\text{discharge}}\},& \Vert\position_t^{\worker{i}} - \position_t^{\station{j}}\Vert > \varepsilon \\
\text{min}\{\capacity{i},\; e_{t-1}^i + e_{\text{charge}}\},& \Vert\position_t^{\worker{i}} - \position_t^{\station{j}}\Vert \leq \varepsilon
\end{cases}
\end{equation}

\subsubsection{Coverage Task and Synchronized Coverage Area}
coverage by \textit{workers} is only carried out when \textit{worker} has energy left. 
Once a \textit{worker} starts to recharge, it would be released from \textit{station} if and only if it is fully recharged (i.e., $p_t^i=1$).
Denote the coverage area of each \textit{worker} $\worker{i}$ at time $t$ as $\coverage_t^i$.
Here we assume the overall covered area $\coverage_t$ at time $t$ can be updated and synchronized among all agents during the task.
The update and synchronization of $\coverage_t$ can be implemented by mutual information exchange for robots within the communication range: $\coverage_t = \bigcup_{t=0}^{T} \bigcup_{i=1}^{m} \coverage_{t}^{i}$.
\begin{figure}[!htp]
\vspace{-1mm}
\centering
\includegraphics[width=0.7\linewidth]{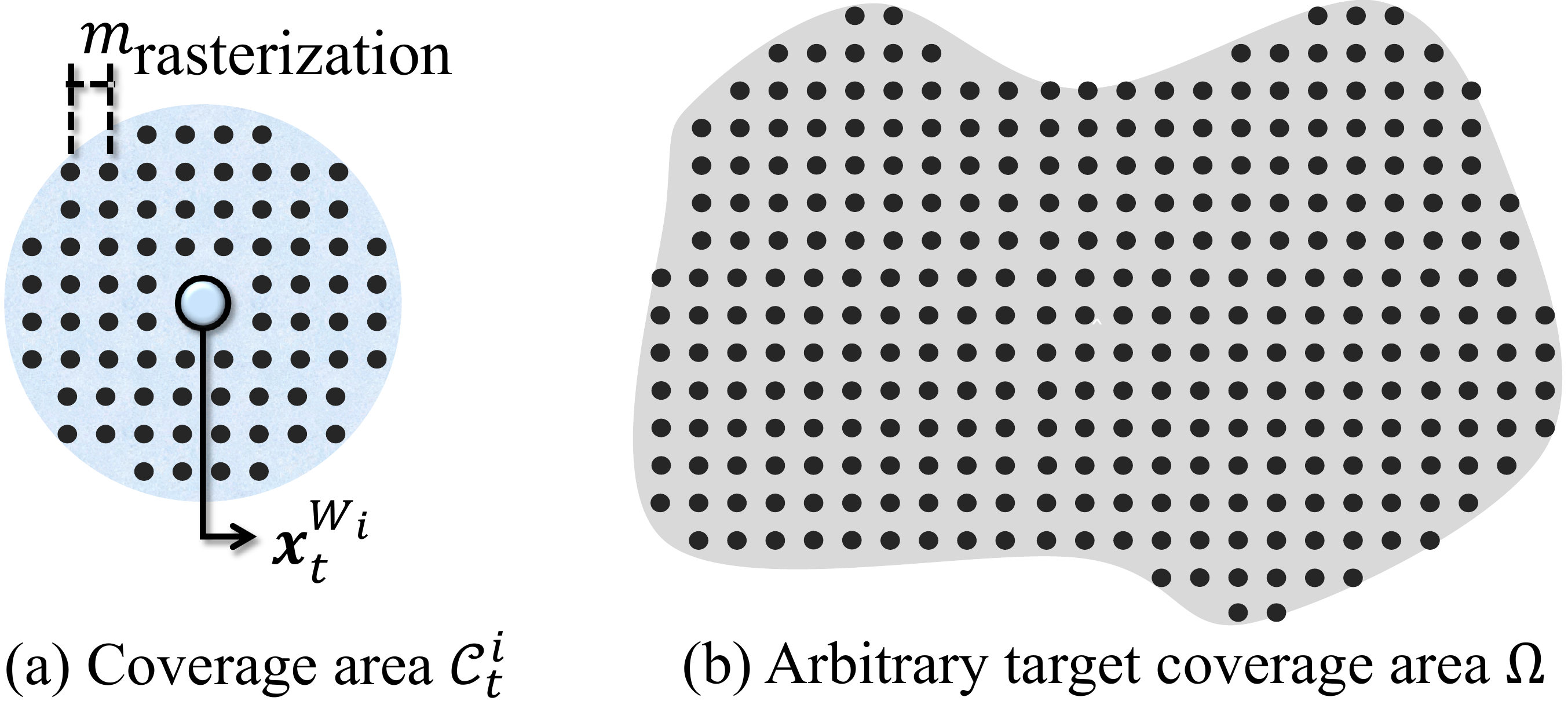}
\caption{Model the coverage task by uniform sampling}
\label{fig:coverage_task_modeling}
\vspace{-1.5mm}
\end{figure}

For a released \textit{worker} $W^i$ with energy left (i.e., $p_t^i>0$), the coverage area $\coverage_t^i$ at time $t$ is determined by uniformly sampling as depicted in Fig.~\ref{fig:coverage_task_modeling}:
given the sampling resolution $m_\text{rasterization}$, the coverage area $\coverage_t^i$ is approximated by uniformly sampling the coordinates within the shape boundaries of the target coverage area $\Omega$ (i.e., rasterization).
Then, the coverage area is represented by a set of coordinates in planar space.
Thus, the union operations on coverage areas $\coverage_t$ turn into set union operations, which is computationally tractable using a hash-set compared with area union operations.


\subsection{Multi-agent Reinforcement Learning}
We first introduce the Decentralized Partially Observable Markov Decision Process (Dec-POMDP)~\cite{oliehoek2016concise} denoted by $\left(\mathcal{R}, \mathcal{S}, \mathcal{A}, \mathcal{P}, \mathcal{O}, r, b \right)$, where $\mathcal{R}$ is the set of agents, $\mathcal{S}$ is the joint state space, $\mathcal{A}$ is the joint action space, $\mathcal{P}$ is the state-transition model, $\mathcal{O}$ is joint observation space, $r$ is the shared reward function and $b$ is the initial state distribution.
With a shared reward function $r$, we can formulate the MARL problem by single-agent reinforcement learning objective~\cite{zhang2021multi}:
\begin{equation}
\begin{split}\label{eq:marl}
\max_\pi \;\mathbb{E}\left[\sum_{t\leq T, s_0 \sim b} \gamma^{t} r\left(s_{t}, a_{t}, s_{t+1}\right)\mid a^i_t\sim\pi(\cdot\mid o^i_t)\right]
\end{split}
\end{equation}
where $\pi$ is the policy and $\gamma$ is reward discount factor, $s_t\in \mathcal{S}$ and $a_t\in\mathcal{A}$ are the joint state and action of agents at time $t$ respectively.
Given the initial state of agents $s_0$ and the observation $o^i_t$ at time $t$, the action $a^i_t$ is sampled from the policy $\pi$.
The goal is to maximize the expected discounted reward within a time horizon of $T$.
Note that Eq.~\ref{eq:marl} can be adopted to agents with different observations or functionalities (each corresponds to a different policy), as long as they share a common reward function.

\subsection{\textit{Worker-Station} MRS Coverage Task Formulation}
As introduced previously, the \textit{worker-station} MRS consists of two types of agents with different functionality: the \textit{workers} are responsible for coverage work with limited energy, whereas the \textit{station} is responsible for replenishing \textit{workers} with unlimited energy.
Thus following Eq.~\ref{eq:marl}, we define the agents $\mathcal{R}=\{W^i\}_{i=1}^m \bigcup \{S^j\}_{j=1}^n$ as the set of \textit{stations} and \textit{workers}, and $\mathcal{A}=\{a^i\}_{i=1}^{m+n}$ are the actions of \textit{workers} and \textit{stations} sequentially.
We define policy $\pi_\phi$ and policy $\pi_\theta$ for \textit{stations} and \textit{workers} respectively, and a shared reward function $r$ for both agents.
Then, we can formalize the mCPP problem for \textit{worker-station} MRS as a fully cooperative MARL problem~\cite{busoniu2008comprehensive}:
\begin{equation}\label{eq:marl_mcpp}
\pi^*_\theta\;, \pi^*_\phi = \argmax_{\pi_\theta, \pi_\phi}\;\mathbb{E}\left[\sum_{t\leq \text{min}\{T, T_{\text{finish}}\}} \gamma^{t} r\left(s_{t}, a_{t}, s_{t+1}\right)\right]
\end{equation}
where $a^i_t\sim\pi_\theta(\cdot\mid o^i_t), i=1,2,...,m$ are the actions of \textit{workers}, and $a^j_t\sim\pi_\phi(\cdot\mid o^j_t), j=m+1,m+2,...,m+n$ are the actions of \textit{stations}.
Note that the planning horizon $T$ is replaced by the minimum of the original time horizon $T$ and the coverage task finish time $T_\text{finish}$.

In order to finish the cooperative coverage task as soon as possible, with collision avoidance and rendezvous to recharge, we define the shared reward function $r$ as below:
\begin{equation}\label{eq:worker_reward}
r = \sum_{i=0}^k\reward{c} + \sum_{i=0}^k\reward{e} + r_{\text{collision}} + r_{\text{time}}
\end{equation}

The first component $\reward{c}$ is the covering reward for $i$-th \textit{worker}, which is the only positive term to guide the coverage planning of \textit{workers}.
The second component $\reward{e}$ is a penalty term added when a \textit{worker}'s energy is close to its energy capacity, which guides the rendezvous planning of \textit{station}.
The details of the first two reward components are elaborated in Sec.~\ref{sec:approach}.
The third component $r_{\text{collision}}$ is a constant collision penalty whenever a collision occurs, which guides the agents for dynamic collision avoidance.
The last component $r_{\text{time}}$ is a constant time penalty in each time step whenever the coverage task has not finished, which helps to find more time-efficient planning policies.

Since the total coverage reward of the coverage area $\Omega$ is a constant (i.e., only new covered area provide rewards), and the other three penalty terms would stop accumulating once the coverage task is finished, only a time-optimal coverage and rendezvous planning policy with collision avoidance can reach the optimal task performance.
Therefore, by designing and selecting appropriate rewards for the above components in Eq.~\ref{eq:marl_mcpp}, we can apply MARL algorithms to train the agents for the \textit{worker-station} MRS coverage task to coordinate with each other for theoretical optimal performance.

\newcommand{\ozero}{^z o}
\newcommand{\operc}{^p o}
\newcommand{\ocomm}{^c o}

\section{Deep Reinforcement Learning Approach} \label{sec:approach}
In this section, we introduce key components of our DRL-based planning method.
We follow the paradigm of centralized training and decentralized execution (CTDE)~\cite{foerster2018counterfactual}, which has been widely used in MARL for Dec-POMDP modeled robot learning problems~\cite{fan2020distributed, tallamraju2020aircaprl}.
The training and planning phases are elaborated in Sec.~\ref{subsec:training_planning}.
In Fig.~\ref{fig:sys_arch}, we summarize our end-to-end planning pipeline.
For an ego agent (\textit{worker} or \textit{station}), the perception-range and communication-range  observations described in Sec.~\ref{subsec:observation} are encoded into feature vectors by a convolution neural network.
Then, they are stacked upon zero-range observation, constituting the final observation vector $\hat{o}_t^i$ in latent space.
The policy network $\pi_{\theta}$ for \textit{worker} and $\pi_{\phi}$ for \textit{station} are both Multi-layer Perceptron (MLP) modules, each of which takes $\hat{o}_t^i$ for the $i$-th agent at time $t$ and output its action $\action_t^i$.
The action $\action_t^i$ is then converted to velocity commands as mentioned in Sec.~\ref{subsec:action}, which solves both rendezvous planning for \textit{station} and coverage planning for \textit{workers}.

\subsection{DRL Training and Planning Phases}\label{subsec:training_planning}
We follow the paradigm of CTDE to train the policy network of each agent in Eq.~\ref{eq:marl_mcpp}, then deploy the corresponding policy networks on robots for planning.
During training, the state of the whole system and observations of all other agents are needed for better training performance in simulation.
During planning, each agent receives only its zero-range, perception-range, and communication-range observations as described in Sec.~\ref{subsec:observation}, then output the best action according to the corresponding trained policy network.
We unfold the details in CTDE in the following two parts.

\subsubsection{Centralized training phase}
for training algorithm, we use the multi-agent actor-critic algorithm MA-POCA~\cite{cohen2021use} to train the policy networks for \textit{workers} and \textit{stations}.
During training, a centralized critic network is trained to estimate the value of the current system state, including the whole system and observations of all agents.
Note that states and observations of the whole system are only needed during training and can be easily accessed in simulations.
According to \cite{lowe2017multi}, such a centralized critic network would greatly help the training of the actor network (i.e., policy network).
\begin{figure}[ht]
\vspace{-2.5mm}
\centering
\includegraphics[width=0.8\linewidth]{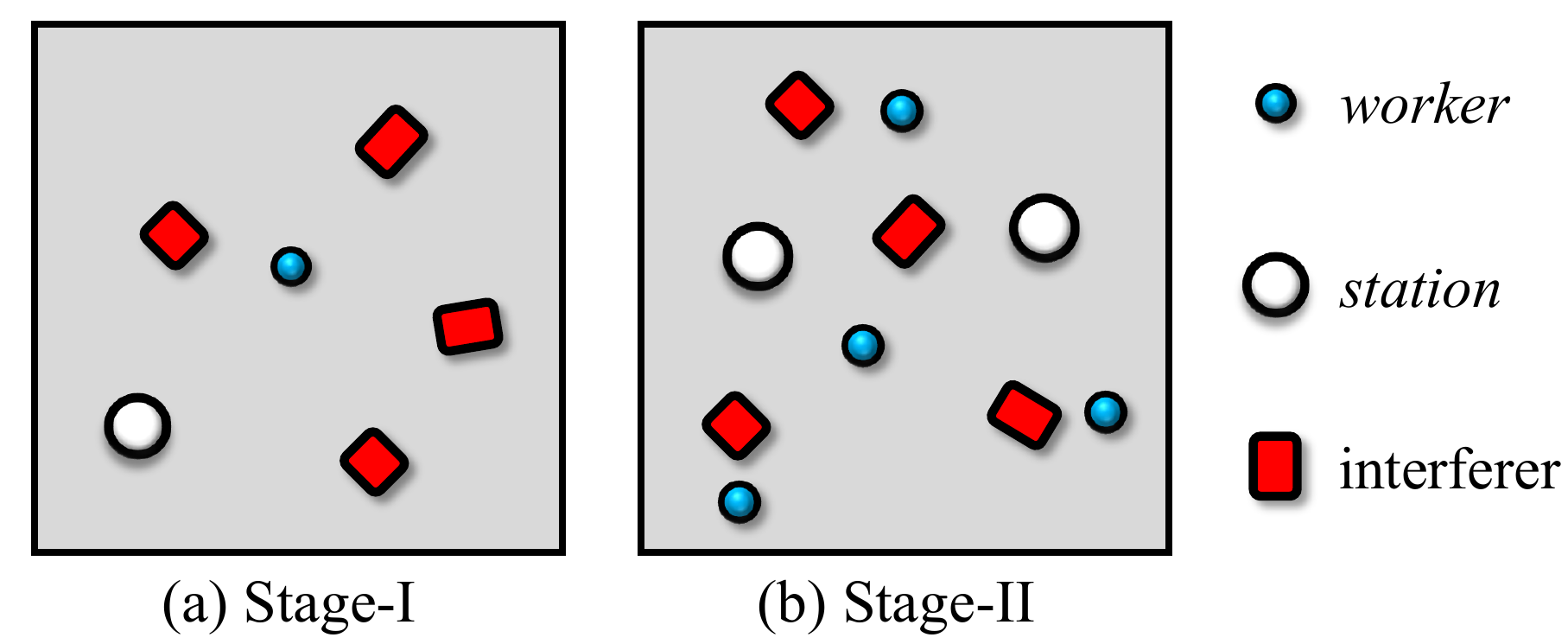}
\caption{Two-stage curriculum learning for mCPP problem of \textit{worker-station} MRS: (a) Stage-I: one \textit{station} with single \textit{worker}; (b) Stage-II: multiple \textit{stations} with multiple \textit{workers}.}
\label{fig:curri_design}
\vspace{-2mm}
\end{figure}

For better policy exploration of the coordination behaviors towards the coverage task during training, we adopt the Intrinsic Curiosity Module (ICM)~\cite{pathak2017curiosity}.
In short, the ICM trains a self-supervised inverse dynamic model that predicts the consequences of an agent's actions, and uses that prediction error as an intrinsic reward to guide the agent's exploration during training.
In considerations of training performance, we designed a two-stage curriculum learning~\cite{bengio2009curriculum} evolving from single \textit{worker} into multiple \textit{workers}, which guides \textit{workers} and \textit{station} for better policies during training.
\begin{figure*}[ht]
\centering
\includegraphics[width=0.90\linewidth]{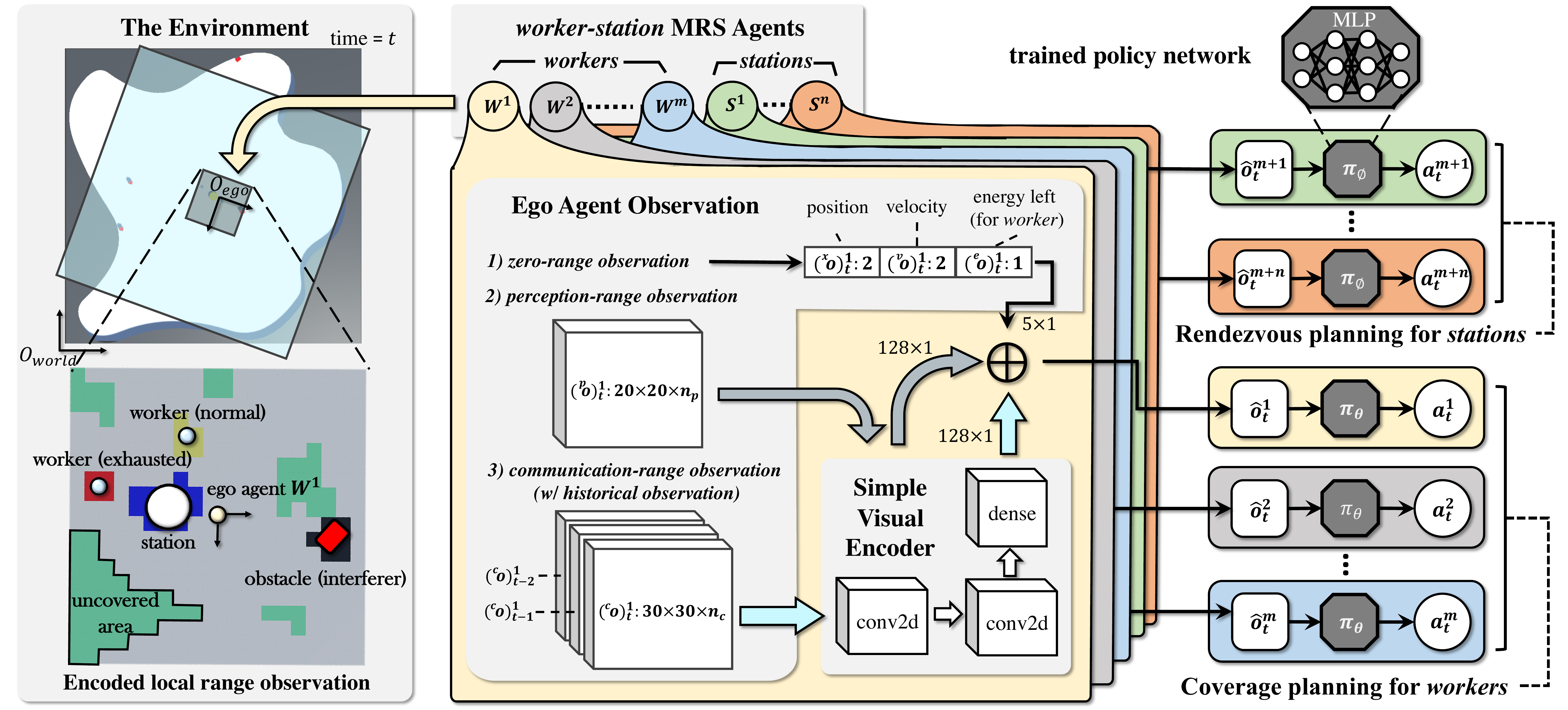}
\caption{Our DRL-based mCPP pipeline for the \textit{worker-station} MRS: during planning, each agent receives its own zero-range, perception-range and communication-range observations, and outputs the best action at each time step according to its trained policy network.}\label{fig:sys_arch}
\vspace{-4.5mm}
\end{figure*}

As shown in Fig.~\ref{fig:curri_design}-(a), stage-I is designed to make it easier for both \textit{worker} and \textit{station} to focus on learning some basic behaviors, such as the ability of collision avoidance with static obstacles and dynamic interferers.
For \textit{worker}, the ``\textit{cover and replenish}'' behavior is learnt when the remained energy of \textit{worker} is at a low level.
For \textit{station}, the behavior of finding and following exhausted \textit{worker} is learnt.
Once training of stage-I is converged, we can then extend the \textit{worker} and \textit{station} to multiple ones, and adapt the pre-trained policy networks to train for final policies (see Fig.~\ref{fig:curri_design}-(b)).

\subsubsection{Decentralized execution phase}
unlike the centralized training phase, each agent only needs its own observation during the decentralized planning phase.
Specifically speaking, each agent only takes its own observation as introduced in \ref{subsec:observation}, and outputs optimal action by its observation and the corresponding policy network $\pi_\phi$ and $\pi_\theta$.

\subsection{Observation Space}\label{subsec:observation}
For both \textit{worker} and \textit{station}, the observation $\hat{o}_t^i$ of the $i$-th ego agent at time $t$ consists of following three types:
1) zero-range observation $(\ozero)_t^i$ contains its own basic information; 2) perception-range observation $(\operc)_t^i$ contains precise local information within its perception range; 3) communication-range observation $(\ocomm)_t^i$ contains rough global information within its communication range.
A demonstration of observation is shown in the ego agent observation block in Fig.~\ref{fig:sys_arch}.
\begin{table}[ht]
\vspace{-5mm}
\begin{center}\begin{tabular}{||c|c c|c c||}
\hline
& \multicolumn{2}{c|}{\textbf{perception-range}} & \multicolumn{2}{|c||}{\textbf{communication-range}} \\
\hline\hline
\multirow{2}{*}{\textit{\textbf{worker}}} & \textit{worker} & \textit{station} & \textit{worker} & \textit{station} \\
& obstacle & uncovered area & \multicolumn{2}{c||}{uncovered area} \\
\hline
\multirow{2}{*}{\textit{\textbf{station}}} &
obstacle & {\textit{worker}(normal)} & \textit{station} & {\textit{worker}(normal)}\\
& \multicolumn{2}{c|}{\textit{worker}(exhausted)} & \multicolumn{2}{c||}{\textit{worker}(exhausted)} \\
\hline 
\end{tabular}
\end{center}
\caption{Encoded objects in perception-range and communication-range observations for ego agents in the \textit{worker-station} MRS.}
\label{tab:observations}
\vspace{-3mm}
\end{table}

Here we elaborate on each type of observation for an ego agent.
For both \textit{workers} and \textit{stations}, $(\ozero)_t^i$ includes global position and local velocity, which are then stacked vertically as 1-D zero-range observation.
Note that when the $i$-th agent is \textit{worker}, the percentage of remaining energy $p_t^i$ is also included in $(\ozero)_t^i$.
Both $(\operc)_t^i$ and $(\ocomm)_t^i$ are encoded as images with object positions (see Fig.~\ref{fig:sys_arch}), which are translated and rotated with the $i$-th ego agent.
The encoded objects in $(\operc)_t^i$ and $(\ocomm)_t^i$ are listed in Tab.~\ref{tab:observations}.
For $(\operc)_t^i$, it is a $20\times20$ image with $n_p$ channels (i.e., the number of encoded objects) and $m_\text{perc}$ grid resolution (i.e., length per pixel).
For $(\ocomm)_t^i$, it is a $30\times30$ image with $n_c$ channels and $m_\text{comm}$ grid resolution.

\subsection{Action Space}\label{subsec:action}
We define the action space as a 2d continuous vector space consisting of linear and angular velocities.
Given the max linear velocity $v^i_\text{max}$ and max angular velocity $\omega^i_\text{max}$ of the $i$-th robot , the sampled action $\action_t^i$ is scaled by multiplying $v^i_\text{max}$ or $\omega^i_\text{max}$ to give the desired velocity commands.

\subsection{Reward Design}
As mentioned in Eq.~\ref{eq:worker_reward}, the shared reward $r$ for all agents consists of four components.
Here we only elaborate on the first two terms since the last two are simply penalty constants as introduced previously.
Recall that the first component $\reward{c}$ is the covering reward for $i$-th \textit{worker} at each time $t$, where positive rewards are given when a new area is covered.
Once the coverage work is completed, the training episode will terminate with a completion reward $r_{\text{finish}}$:
\begin{equation}
\reward{c} = \begin{cases}
r_{\text{finish}},\qquad\qquad\quad \Omega=\bigcup_{t'=0}^t \bigcup_{i=1}^k \coverage_{t'}^i\\[1pt]
r_{\text{cover}}\times\left(\lvert\coverage_{t}^i\rvert - \lvert\coverage_{t-1}^i\rvert\right),\quad\;\;\; \text{otherwise}
\end{cases}
\end{equation}

The second component $\reward{e}$ is a soft approximation modeling on the hard constraint of \textit{worker}'s capacity.
For \textit{worker} $\worker{i}$, it allows $p_t^i$ to be less than zero during training, which let $\worker{i}$ still be able to move when $p_t^i\leq0$ (i.e., no energy left).
More specifically, such a soft approximation uses a truncated exponential function for $\reward{e}$ as below:
\begin{equation}
\reward{e} = \begin{cases}
-1\times \text{min}\{1, \exp\,(p_t^i-p^e)\},& p_t^i < p^e\\[1pt]
0,& \text{otherwise}
\end{cases}
\end{equation}
where $p^e$ is the threshold indicating whether the \textit{worker} is exhausted.
Such design results from practical considerations:
1) direct modeling such hard constraint during training makes \textit{worker} struggles to learn the ``\textit{cover and replenish}" behavior when energy is exhausted;
2) the truncated exponential penalty approximation with a relatively large derivative around $p^e$ makes \textit{worker} aware of its exhausted status when $p_t^i$ approaches $p^e$.
Note that the energy capacity hard constraint of \textit{worker} is only modeled as a soft constraint during training; it is still a hard constraint (i.e., \textit{workers} cannot move once $p_t^i \leq 0$) during planning.

\section{Experiments \& Results} \label{sec:exp}
\subsection{Implementation Details}
Since we mainly focus on strategy-level planning problems in this paper, we use Unity and ML-Agents toolkit~\cite{juliani2020unity} to build the environment and system.
The dynamic interferers are modeled in a loop to first move in a constant speed and a random direction within a given period, and then rotate with a random angle.
Such a loop for interferers repeats until the coverage task finishes.
Also, \textit{worker} can only be replenished when it is exhausted (i.e., $p_t^i<p^e$) and near the \textit{station}.


\subsection{Simulation Results}
\begin{figure}[h!t]
\vspace{-3mm}
\centering
\includegraphics[width=\linewidth]{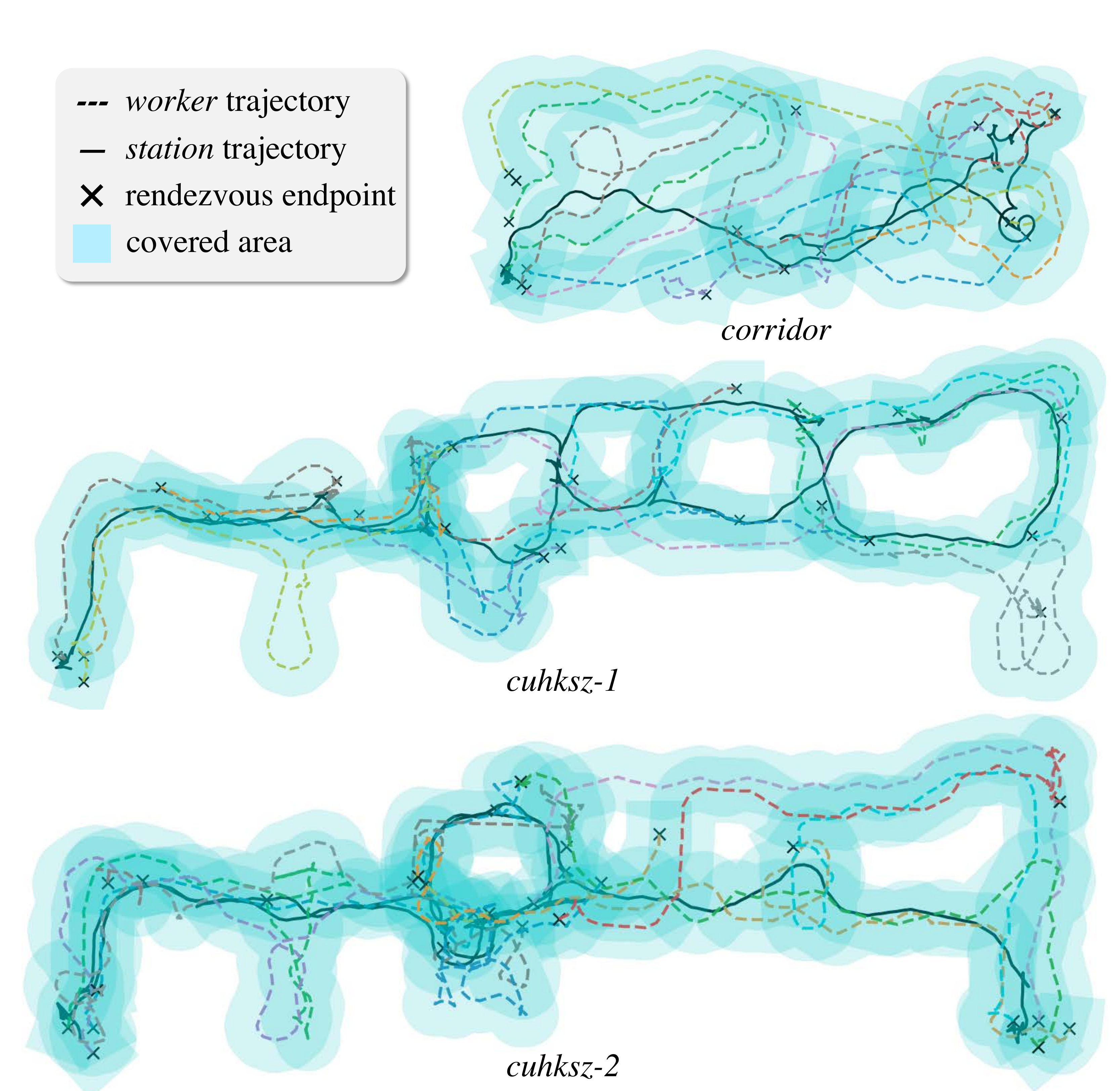}
\caption{motion trajectories of \textit{worker-station} MRS using our method.}
\label{fig:sim_traj}
\vspace{-3mm}
\end{figure}

We modeled three simulation scenes in Unity to conduct simulation experiments, including ablation study and the coverage task performance comparison.
\begin{figure}[h!t]
\vspace{-3mm}
\centering
\includegraphics[width=0.9\linewidth]{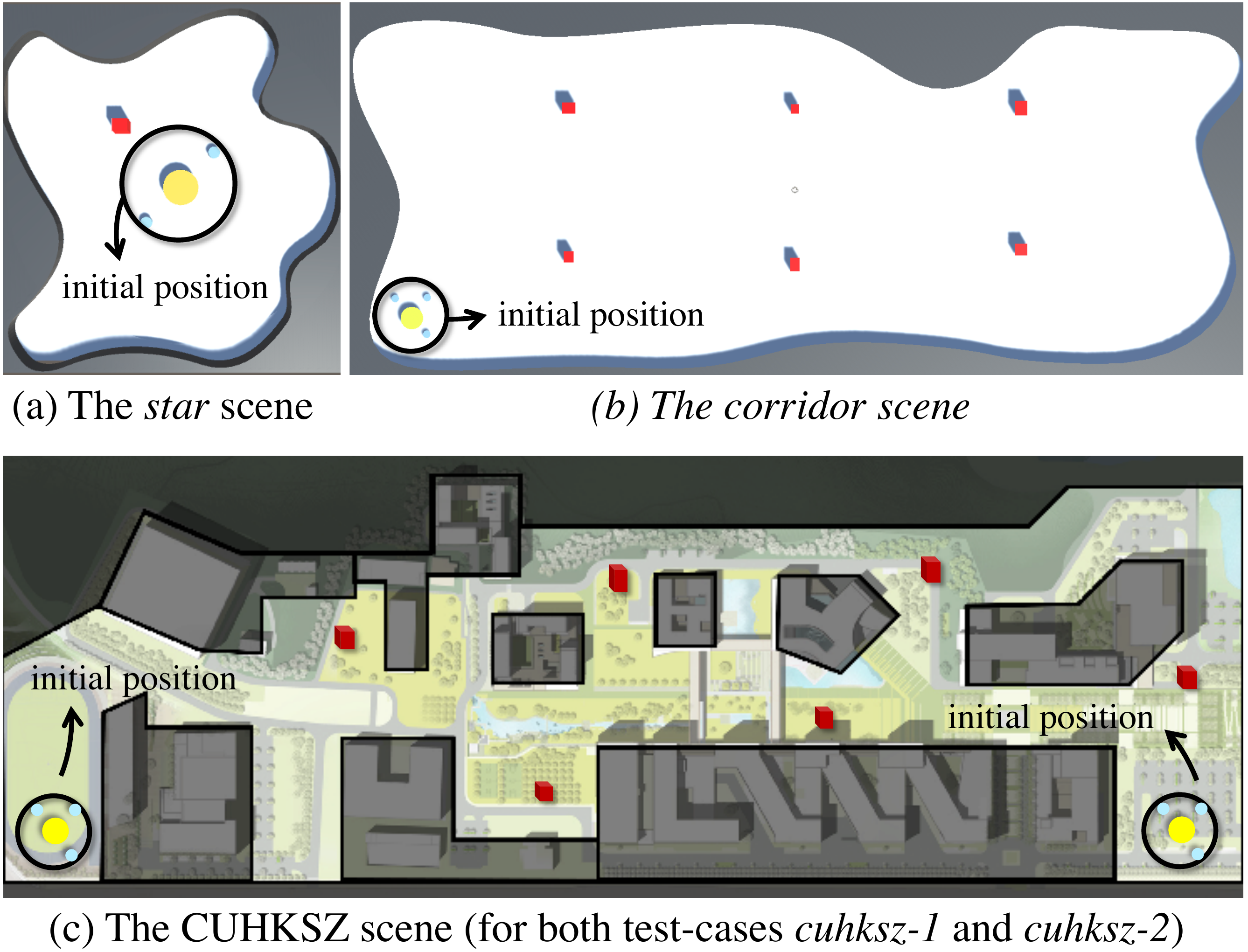}
\caption{Modeled simulation scenes in Unity. The target coverage areas are bounded within the grey obstacle areas.}
\label{fig:sim_scene}
\vspace{-5.5mm}
\end{figure}
\begin{table}[ht]
\begin{center}\begin{tabular}{||c||c|c|c|c||}
\hline
\textbf{test-case name} & \textbf{\textit{star}} & \textbf{\textit{corridor}} & \textbf{\textit{cuhksz-1}} & \textbf{\textit{cuhksz-2}} \\
\hline
\textbf{target area size} & 30$\times$30 & 120$\times$50 & 180$\times$60 & 180$\times$60\\
\hline
\textbf{\textit{worker} cover radius} & 4 & 4 & 2 & 2\\
\hline
\textbf{\# of \textit{workers}} & 2 & 3 & 3 & 6\\
\hline
\textbf{\# of \textit{stations}} & 1 & 1 & 1 & 2\\
\hline
\textbf{\# of \textit{interferers}} & 1 & 6 & 6 & 6 \\
\hline
\end{tabular}
\end{center}
\caption{Design details of simulation test-cases.}
\label{tab:design_sim_scene}
\vspace{-3mm}
\end{table}
As in Fig.~\ref{fig:sim_scene}-(a) and (b), two irregularly shaped scenes are used in simulation experiments, where robots of the \textit{worker-station} MRS are initialized in the initial position.
In addition, we modeled the CUHKSZ campus for coverage work in Fig.~\ref{fig:sim_scene}-(c), where the buildings are considered static obstacles.
Fig.~\ref{fig:sim_traj} shows the motion trajectories of the \textit{worker-station} using our planning method in simulation test-cases.
Based on the three modeled scenes, we designed four test-cases described in Tab.~\ref{tab:design_sim_scene}.
Note that in \textit{cuhksz-2} test-case, only the left-bottom group of robots is included for the coverage work.

\subsubsection{Ablation Study}
\begin{figure}[h!t]
\vspace{-3mm}
\centering
\includegraphics[width=0.97\linewidth]{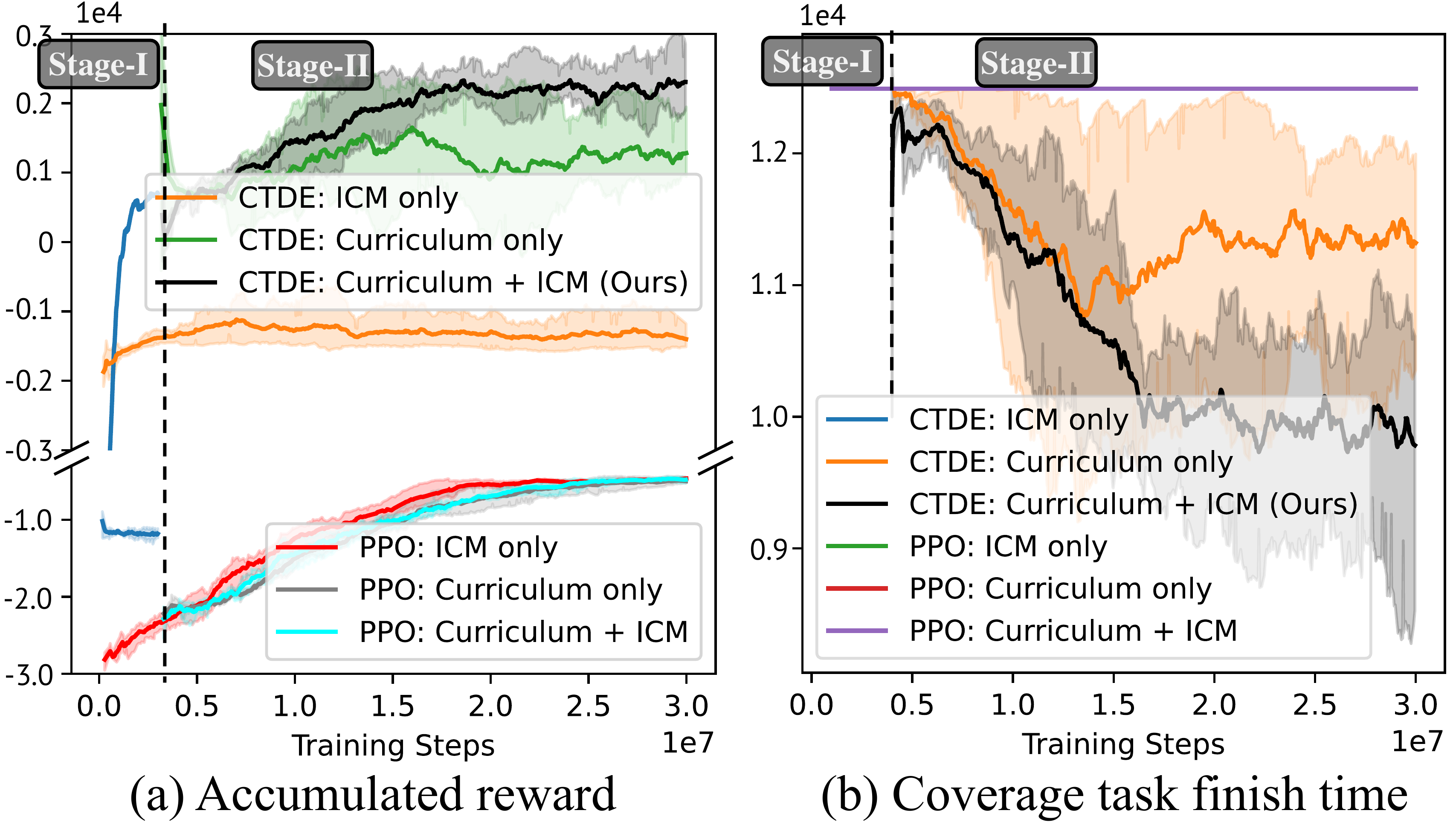}
\caption{Ablation study of two-stage curriculum learning, Intrinsic Curiosity Module (ICM) and centralized PPO in \textit{corridor} test-case.}
\label{fig:ablation_study}
\vspace{-2.5mm}
\end{figure}

to validate the effects on training performance brought by our curriculum learning design and ICM, we conducted ablation study in the \textit{corridor} test-case.
We also trained a centralized policy with PPO~\cite{schulman2017proximal} to validate the benefits of the CTDE decentralization paradigm.
In short, the PPO agent takes the observation of object-positions encoded images, which should cover the whole coverage area with high resolution as in the perception-range observation.
Therefore, it is much larger than the image observations for each ego agent in CTDE.
With the same visual encoder in Fig.~\ref{fig:sys_arch}, the feature vectors are fed into MLP of the same size to output the joint actions that are distributed to each robot.
It is evident in Fig.~\ref{fig:ablation_study} that a centralized policy using PPO failed in our problem.
Such a failure is largely due to: 1) the training difficulties on a much larger network (about $1.5e6$ parameters with centralized PPO, $0.1e6$ parameters for both \textit{worker} and \textit{station} policies with CTDE); 2) and the lack of cooperation between agents with centralized PPO.

We now compare the results within the CTDE paradigm.
For two-stage curriculum learning, as shown in Fig.~\ref{fig:ablation_study}, when training from scratch without curriculum, agents in the \textit{worker-station} MRS struggle at a locally optimal policy and cannot finish the coverage task.
When initializing from Stage-I, it provides basic policy networks for both \textit{worker} and \textit{station}, which vastly improves the sample efficiency and guides the training procedure.
As for ICM, we first initialize policy networks from a pre-trained Stage-I curriculum learning.
As shown in Fig.~\ref{fig:ablation_study}-(b), the task finish time $t_\text{finish}$ shows that agents trained with ICM are better than agents trained without it, which reflects the reward gap between two training curves (green and black) in Fig.~\ref{fig:ablation_study}-(a).
Such reward gap results from the earlier finish of the coverage task, which eliminates more accumulating time penalty $r_\text{time}$.

\subsubsection{Decomposition-based and Graph-based Baselines}
to evaluate the coverage task performance, we modified graph-based and decomposition-based mCPP methods to several heuristic baseline methods on discretized state space.
In addition, since these offline centralized mCPP baseline methods have no dynamic collision avoidance ability with interferers, we adopt a \textit{wait-and-move} policy for all baseline methods.

\textbf{Mobile-BCD:} for decomposition-based baseline method, we follow the Boustrophedon Cellular Decomposition (BCD) algorithm~\cite{choset2000coverage} and adopts it as the so-called \textit{mobile-BCD} for our problem.
We briefly describe the procedure: 1) the map is initially decomposed into cells via BCD; 2) in each cell, back-and-forth trajectories on the uncovered area are generated and evenly distributed to \textit{workers}; 3) the \textit{stations} always move to the nearest exhausted \textit{worker} to replenish it.

\textbf{Static-MSTC*:} for graph-based mCPP baseline methods, we first modify the state-of-the-art mCPP algorithm MSTC*~\cite{tang2021mstc} into a static \textit{stations} version, namely \textit{static-MSTC*}.
In order to account for the continuous energy capacity of \textit{workers} in our problem setting, the key modification in static-MSTC* is the constraint approximation from the node-based energy capacity constraint in the original MSTC* to the travel time-based constraint as in Eq.~\ref{eq:energy_model}.

\textbf{Mobile-MSTC*:} based on the static-MSTC* method, we further mobilize the \textit{stations} and design the \textit{mobile-MSTC*} baseline as follows:
1) the target area is first decomposed into sub-regions via \textit{k-means clustering};
2) depth-first-search is applied to plan for the \textit{stations} loaded with \textit{workers}, to travel to the center of next uncovered sub-region;
3) at each sub-region, the \textit{workers} cover the area via the static-MSTC* baseline method.
Note that the $k$ value in the \textit{k-means clustering} algorithm is chosen according to the capacity $\capacity{i}$ to make partitions suitable for efficient planning.

\subsubsection{Coverage Task Performance}
we compared the coverage task finish time $T_\text{finish}$ among our method and the above baseline methods on all the test-cases in Tab.~\ref{tab:design_sim_scene}.
The smaller $T_\text{finish}$ is, the better the planning strategy for the mCPP problem is.
Note that to adopt the mobile-MSTC* baseline method, we decompose the CUHKSZ map into two equal-sized areas, and each area runs the mobile-MSTC* baseline separately to finish the coverage work.

\textbf{Test-cases \textit{star}, \textit{corridor}, \textit{cuhksz-2}:}
we first compare static-MSTC* with mobile-MSTC* and mobile-BCD.
In test-cases \textit{star} and \textit{corridor}, the performance of mobile-MSTC* and mobile-BCD is nearly the same as static-MSTC*.
In test-case \textit{cuhksz}-2, the mobile-MSTC* and mobile-BCD manage to improve task performance by mobilizing the \textit{station} and planning for \textit{station} and \textit{workers} separately.
However, there remains vast space for coverage task performance improvement; the mobile-station and mobile-BCD baselines that separately plan for \textit{stations} and \textit{workers} with dynamic interferers still perform inefficiently.

We now compare our method with baseline methods.
In general, the comparison results on the three test-cases show that our planning method can generate good coordination behaviors of coverage planning and rendezvous planning for \textit{workers} and \textit{station}, which leads to a better performance in metrics of $t_\text{finish}$ after around $0.5e7$ to $1.5e7$ training steps.
Compared with the mobile-MSTC* baseline in test-cases \textit{corridor} and \textit{cuhksz-2} with larger target areas, our method unlocks more benefits by mobilizing the \textit{station} and utilizing the mobility of each robot in the MRS.
Interestingly, when an exhausted \textit{worker} leaves the perception or communication range of \textit{station}, the rendezvous for recharge is still possible, as \textit{stations} would explore to search exhausted \textit{workers}.
\begin{figure}[h!tp]
\vspace{-3mm}
\centering
\includegraphics[width=0.97\linewidth]{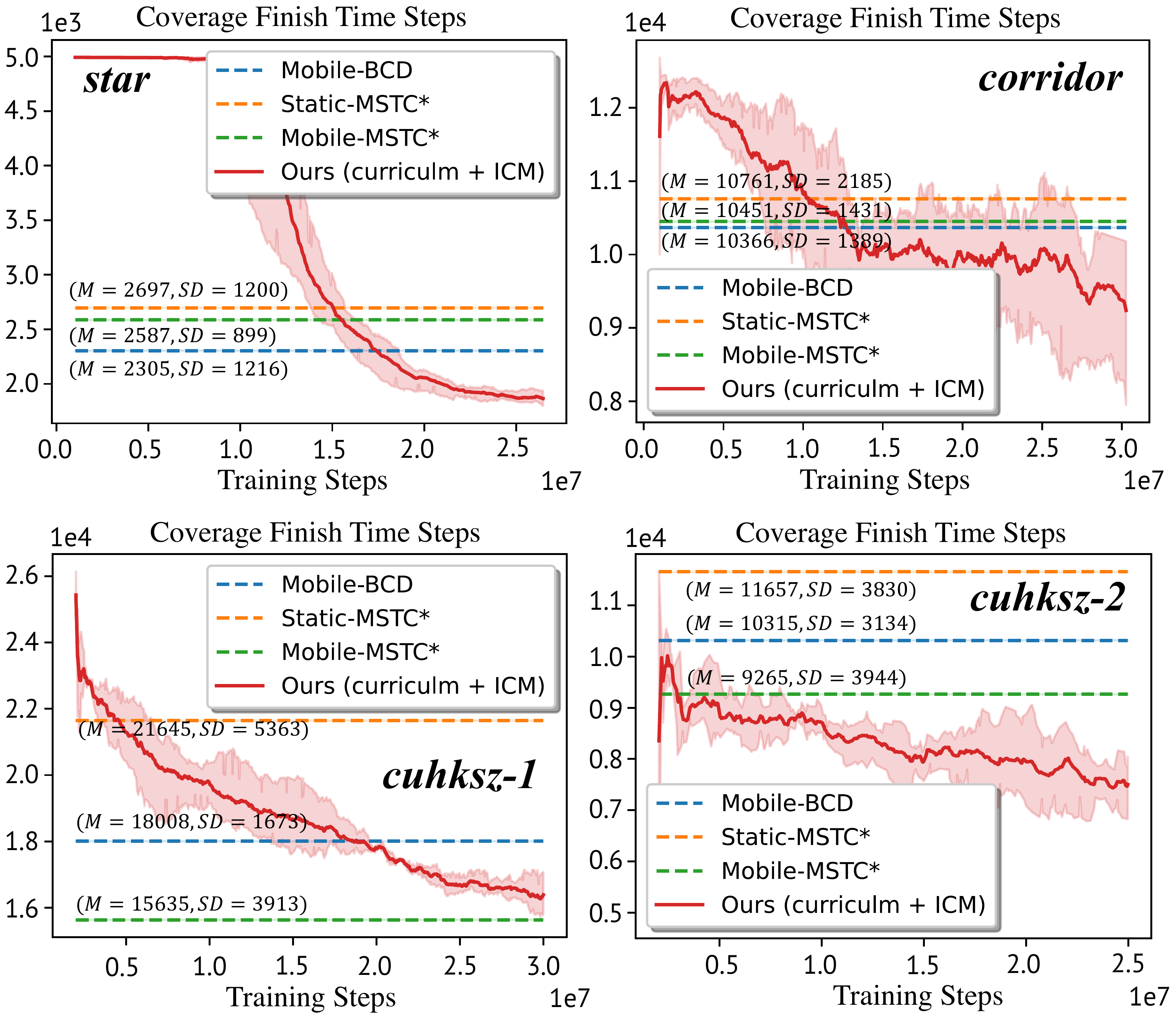}
\caption{Comparisons of the coverage task finish time.}
\label{fig:cov_task_performance}
\vspace{-3mm}
\end{figure}

\textbf{Test-case \textit{cuhksz-1}:}
compared with test-case \textit{cuhksz-2}, the MRS of test-case \textit{cuhksz-1} works in the same CUHKSZ scene but with the numbers of \textit{workers} and \textit{stations} reduced in half.
The performance of our method is nearly the same as mobile-MSTC* with the best performance, which is mainly due to the following reason:
for less number of \textit{workers} and a comparably smaller cover radius, \textit{workers} with continuous action space would leave uncovered gaps during work, especially when trying to avoid static obstacles or dynamic interferers.
These uncovered gaps require the \textit{workers} to revisit some regions, making our method in this test-case perform not as well as in the other three test-cases.

\subsection{Real Robot Performance}
\begin{figure}[htp]
\vspace{-2.5mm}
\centering
\includegraphics[width=0.9\linewidth]{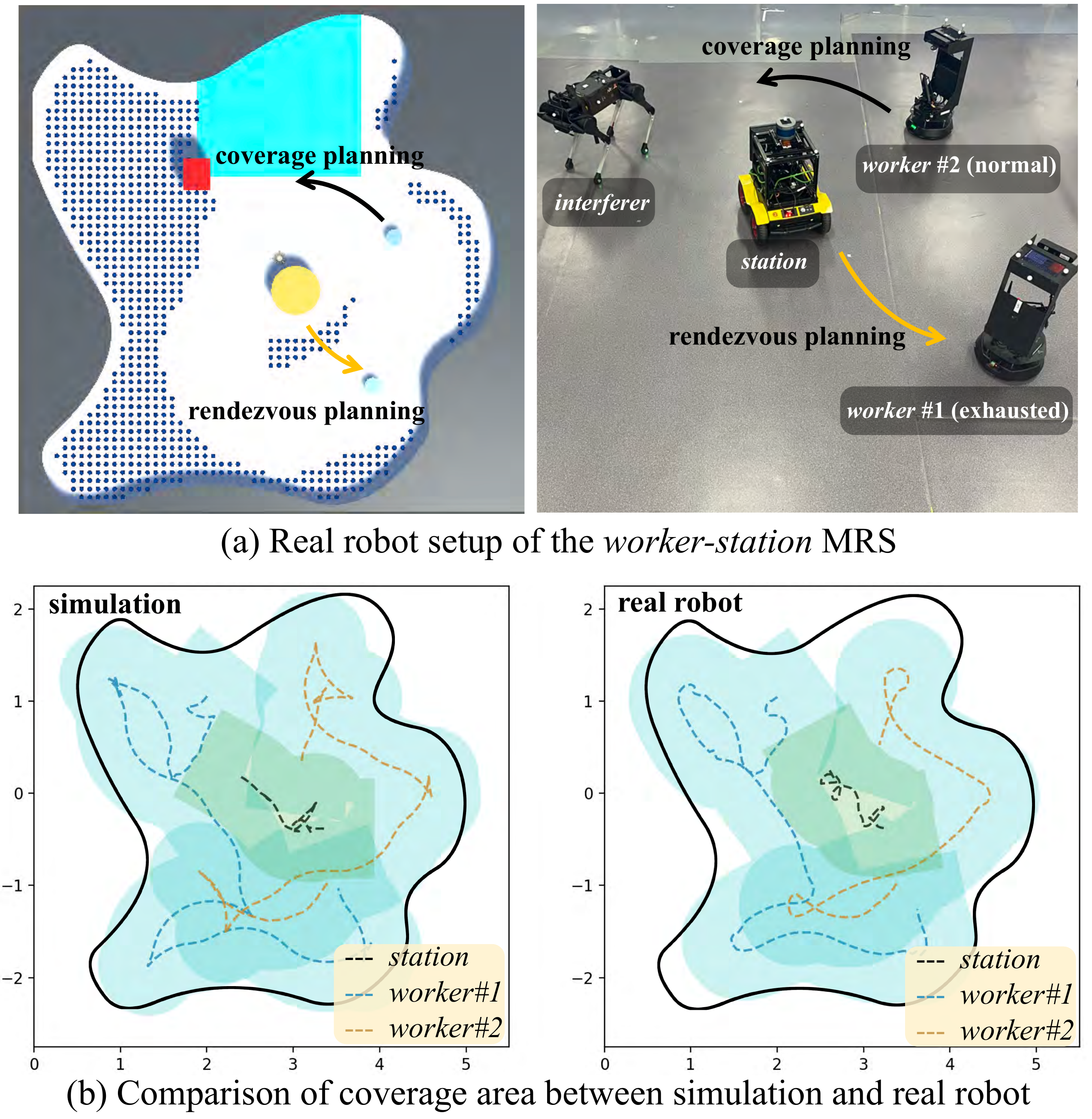}
\caption{Real robot demonstration of our planning method.}
\label{fig:realexp}
\vspace{-2mm}
\end{figure}

As complementary to the simulated environments, we also conduct hardware experiments of our method on real robots.
As shown in Fig.~\ref{fig:realexp}, we tested our method in \textit{Star} scene, of which we made a replica in the real world.
The \textit{worker-station} MRS consists of two \textit{workers} (black differential-driven wheeled robots) and one \textit{station} (yellow skid-steer wheeled robot), the dynamic interferer is a quadruped robot.
The \textit{workers} are considered replenished once its distance to the \textit{station} is smaller than a threshold.
We use the PID controller for velocity commands of all robots, with a motion capture system providing their global position information.

Fig.~\ref{fig:realexp}-(a) depicts the \textit{worker-station} MRS with our method, where the \textit{station} is moving towards \textit{worker} \#2 to replenish it and \textit{worker} \#1 is executing coverage work.
Fig.~\ref{fig:realexp}-(b) is a comparison of the coverage area between simulation and real robot, the blue area is covered by \textit{workers} and the green area is the motion range of \textit{station}, the whole coverage task in the real world took $140$ seconds.

\subsection{Discussions}
To the best of our knowledge, there is no existing method for online and simultaneous planning for the \textit{worker-station} MRS.
Therefore, our choice of the baseline methods is naive heuristics-based planning approaches, which also need extensive research to reach optimal performance with fine-tuned hyperparameters.
There are several limitations of our DRL-based planning method.
First, when there is only a comparably small number of \textit{workers} with a small cover radius, the unregulated trajectories of \textit{workers} with continuous action space would leave more uncovered gaps needed for revisiting.
Second, since our method mainly focuses on strategy-level planning for \textit{workers} and \textit{station} towards the coverage task, we use a relatively simple controller for the generated velocity actions to control real robots, which could cause a performance gap between simulation and reality.

\section{Conclusions \& Future Work}
In this paper, we introduce the \textit{worker-station} Multi-robot System (MRS) to solve the Multi-robot Coverage Path Planning (mCPP) problem, which can be generalized to various applications in the real world.
We provide a fully cooperative multi-agent reinforcement learning formulation of the above problem, and propose an end-to-end decentralized online planning method based on Deep Reinforcement Learning.
Our method simultaneously plans for \textit{workers} and \textit{station} to work together and utilize the mobility of each robot toward the coverage task.
We conduct ablation study, simulation and real robot experiments and demonstrations.
The experimental results show that our method is more efficient in planning for \textit{workers} and \textit{station}, and our method can better utilize the mobility of each robot compared with the mobile-station baseline method.
For future work, there are two directions to further improve the coverage task performance based on our method.
First, by regulating the trajectories with the graph-based mCPP method, there would be fewer missing gaps after \textit{workers} covered an area.
However, it might potentially raise the time for random dynamic collision avoidance.
Second, by explicitly pre-allocating or negotiating which exhausted \textit{workers} should the \textit{stations} be responsible for, the \textit{stations} can be more efficient to replenish specific \textit{workers}.

\addtolength{\textheight}{0cm}   

\bibliographystyle{IEEEtran}
\bibliography{ref}

\begin{thebibliography}{10}
\providecommand{\url}[1]{#1}
\csname url@rmstyle\endcsname
\providecommand{\newblock}{\relax}
\providecommand{\bibinfo}[2]{#2}
\providecommand\BIBentrySTDinterwordspacing{\spaceskip=0pt\relax}
\providecommand\BIBentryALTinterwordstretchfactor{4}
\providecommand\BIBentryALTinterwordspacing{\spaceskip=\fontdimen2\font plus
\BIBentryALTinterwordstretchfactor\fontdimen3\font minus
  \fontdimen4\font\relax}
\providecommand\BIBforeignlanguage[2]{{%
\expandafter\ifx\csname l@#1\endcsname\relax
\typeout{** WARNING: IEEEtran.bst: No hyphenation pattern has been}%
\typeout{** loaded for the language `#1'. Using the pattern for}%
\typeout{** the default language instead.}%
\else
\language=\csname l@#1\endcsname
\fi
#2}}

\bibitem{liu2016multirobot}
Y.~Liu and G.~Nejat, ``Multirobot cooperative learning for semiautonomous
  control in urban search and rescue applications,'' \emph{Journal of Field
  Robotics}, vol.~33, no.~4, pp. 512--536, 2016.

\bibitem{palacios2016distributed}
J.~M. Palacios-Gas{\'o}s, E.~Montijano, C.~Sag{\"u}{\'e}s, and S.~Llorente,
  ``Distributed coverage estimation and control for multirobot persistent
  tasks,'' \emph{IEEE transactions on Robotics}, vol.~32, no.~6, pp.
  1444--1460, 2016.

\bibitem{schuster2020arches}
M.~J. Schuster, M.~G. M{\"u}ller, S.~G. Brunner, H.~Lehner, P.~Lehner,
  R.~Sakagami, A.~D{\"o}mel, L.~Meyer, B.~Vodermayer, R.~Giubilato,
  \emph{et~al.}, ``The arches space-analogue demonstration mission: towards
  heterogeneous teams of autonomous robots for collaborative scientific
  sampling in planetary exploration,'' \emph{IEEE Robotics and Automation
  Letters}, vol.~5, no.~4, pp. 5315--5322, 2020.

\bibitem{couture2009adaptive}
A.~Couture-Beil and R.~T. Vaughan, ``Adaptive mobile charging stations for
  multi-robot systems,'' in \emph{2009 IEEE/RSJ international conference on
  intelligent robots and systems}.\hskip 1em plus 0.5em minus 0.4em\relax IEEE,
  2009, pp. 1363--1368.

\bibitem{litus2007frugal}
Y.~Litus, R.~T. Vaughan, and P.~Zebrowski, ``The frugal feeding problem:
  Energy-efficient, multi-robot, multi-place rendezvous,'' in \emph{Proceedings
  2007 IEEE International Conference on Robotics and Automation}.\hskip 1em
  plus 0.5em minus 0.4em\relax IEEE, 2007, pp. 27--32.

\bibitem{almadhoun2019survey}
R.~Almadhoun, T.~Taha, L.~Seneviratne, and Y.~Zweiri, ``A survey on multi-robot
  coverage path planning for model reconstruction and mapping,'' \emph{SN
  Applied Sciences}, vol.~1, no.~8, pp. 1--24, 2019.

\bibitem{litus2008distributed}
Y.~Litus, P.~Zebrowski, and R.~T. Vaughan, ``A distributed heuristic for
  energy-efficient multirobot multiplace rendezvous,'' \emph{IEEE Transactions
  on Robotics}, vol.~25, no.~1, pp. 130--135, 2008.

\bibitem{wang2017safety}
L.~Wang, A.~D. Ames, and M.~Egerstedt, ``Safety barrier certificates for
  collisions-free multirobot systems,'' \emph{IEEE Transactions on Robotics},
  vol.~33, no.~3, pp. 661--674, 2017.

\bibitem{MFC-IROS'05}
{Xiaoming Zheng}, {Sonal Jain}, S.~{Koenig}, and D.~{Kempe}, ``Multi-robot
  forest coverage,'' in \emph{IEEE/RSJ International Conference on Intelligent
  Robots and Systems}, 2005, pp. 3852--3857.

\bibitem{MFC-2010}
X.~{Zheng}, S.~{Koenig}, D.~{Kempe}, and S.~{Jain}, ``Multirobot forest
  coverage for weighted and unweighted terrain,'' \emph{IEEE Transactions on
  Robotics}, vol.~26, no.~6, pp. 1018--1031, 2010.

\bibitem{kapoutsis2017darp}
A.~C. Kapoutsis, S.~A. Chatzichristofis, and E.~B. Kosmatopoulos, ``{DARP}:
  divide areas algorithm for optimal multi-robot coverage path planning,''
  \emph{Journal of Intelligent \& Robotic Systems}, vol.~86, no. 3-4, pp.
  663--680, 2017.

\bibitem{rekleitis2008efficient}
I.~Rekleitis, A.~P. New, E.~S. Rankin, and H.~Choset, ``Efficient boustrophedon
  multi-robot coverage: an algorithmic approach,'' \emph{Annals of Mathematics
  and Artificial Intelligence}, vol.~52, no.~2, pp. 109--142, 2008.

\bibitem{collins2021scalable}
L.~Collins, P.~Ghassemi, E.~T. Esfahani, D.~Doermann, K.~Dantu, and
  S.~Chowdhury, ``Scalable coverage path planning of multi-robot teams for
  monitoring non-convex areas,'' in \emph{2021 IEEE International Conference on
  Robotics and Automation (ICRA)}.\hskip 1em plus 0.5em minus 0.4em\relax IEEE,
  2021, pp. 7393--7399.

\bibitem{azpurua2018multi}
H.~Azp{\'u}rua, G.~M. Freitas, D.~G. Macharet, and M.~F. Campos, ``Multi-robot
  coverage path planning using hexagonal segmentation for geophysical
  surveys,'' \emph{Robotica}, vol.~36, no.~8, pp. 1144--1166, 2018.

\bibitem{sun2021ft}
C.~Sun, J.~Tang, and X.~Zhang, ``{FT-MSTC*}: An efficient fault tolerance
  algorithm for multi-robot coverage path planning,'' in \emph{2021 IEEE
  International Conference on Real-time Computing and Robotics (RCAR)}.\hskip
  1em plus 0.5em minus 0.4em\relax IEEE, 2021, pp. 107--112.

\bibitem{mathew2015multirobot}
N.~Mathew, S.~L. Smith, and S.~L. Waslander, ``Multirobot rendezvous planning
  for recharging in persistent tasks,'' \emph{IEEE Transactions on Robotics},
  vol.~31, no.~1, pp. 128--142, 2015.

\bibitem{yu2018algorithms}
K.~Yu, A.~K. Budhiraja, and P.~Tokekar, ``Algorithms for routing of unmanned
  aerial vehicles with mobile recharging stations,'' in \emph{2018 IEEE
  International Conference on Robotics and Automation (ICRA)}.\hskip 1em plus
  0.5em minus 0.4em\relax IEEE, 2018, pp. 5720--5725.

\bibitem{sun2020unified}
C.~Sun, N.~Kingry, and R.~Dai, ``A unified formulation and nonconvex
  optimization method for mixed-type decision-making of robotic systems,''
  \emph{IEEE Transactions on Robotics}, vol.~37, no.~3, pp. 831--846, 2020.

\bibitem{seyedi2019persistent}
S.~Seyedi, Y.~Yazicio{\u{g}}lu, and D.~Aksaray, ``Persistent surveillance with
  energy-constrained uavs and mobile charging stations,''
  \emph{IFAC-PapersOnLine}, vol.~52, no.~20, pp. 193--198, 2019.

\bibitem{oliehoek2016concise}
F.~A. Oliehoek and C.~Amato, \emph{A concise introduction to decentralized
  POMDPs}.\hskip 1em plus 0.5em minus 0.4em\relax Springer, 2016.

\bibitem{zhang2021multi}
K.~Zhang, Z.~Yang, and T.~Ba{\c{s}}ar, ``Multi-agent reinforcement learning: A
  selective overview of theories and algorithms,'' \emph{Handbook of
  Reinforcement Learning and Control}, pp. 321--384, 2021.

\bibitem{busoniu2008comprehensive}
L.~Busoniu, R.~Babuska, and B.~De~Schutter, ``A comprehensive survey of
  multiagent reinforcement learning,'' \emph{IEEE Transactions on Systems, Man,
  and Cybernetics, Part C (Applications and Reviews)}, vol.~38, no.~2, pp.
  156--172, 2008.

\bibitem{foerster2018counterfactual}
J.~Foerster, G.~Farquhar, T.~Afouras, N.~Nardelli, and S.~Whiteson,
  ``Counterfactual multi-agent policy gradients,'' in \emph{Proceedings of the
  AAAI Conference on Artificial Intelligence}, vol.~32, no.~1, 2018.

\bibitem{fan2020distributed}
T.~Fan, P.~Long, W.~Liu, and J.~Pan, ``Distributed multi-robot collision
  avoidance via deep reinforcement learning for navigation in complex
  scenarios,'' \emph{The International Journal of Robotics Research}, vol.~39,
  no.~7, pp. 856--892, 2020.

\bibitem{tallamraju2020aircaprl}
R.~Tallamraju, N.~Saini, E.~Bonetto, M.~Pabst, Y.~T. Liu, M.~J. Black, and
  A.~Ahmad, ``Aircaprl: autonomous aerial human motion capture using deep
  reinforcement learning,'' \emph{IEEE Robotics and Automation Letters},
  vol.~5, no.~4, pp. 6678--6685, 2020.

\bibitem{cohen2021use}
A.~Cohen, E.~Teng, V.-P. Berges, R.-P. Dong, H.~Henry, M.~Mattar, A.~Zook, and
  S.~Ganguly, ``On the use and misuse of absorbing states in multi-agent
  reinforcement learning,'' \emph{arXiv preprint:2111.05992}, 2021.

\bibitem{lowe2017multi}
R.~Lowe, Y.~Wu, A.~Tamar, J.~Harb, P.~Abbeel, and I.~Mordatch, ``Multi-agent
  actor-critic for mixed cooperative-competitive environments,'' in
  \emph{Proceedings of the 31st International Conference on Neural Information
  Processing Systems}, 2017, pp. 6382--6393.

\bibitem{pathak2017curiosity}
D.~Pathak, P.~Agrawal, A.~A. Efros, and T.~Darrell, ``Curiosity-driven
  exploration by self-supervised prediction,'' in \emph{International
  conference on machine learning}.\hskip 1em plus 0.5em minus 0.4em\relax PMLR,
  2017, pp. 2778--2787.

\bibitem{bengio2009curriculum}
Y.~Bengio, J.~Louradour, R.~Collobert, and J.~Weston, ``Curriculum learning,''
  in \emph{Proceedings of the 26th annual international conference on machine
  learning}, 2009, pp. 41--48.

\bibitem{juliani2020unity}
A.~Juliani, V.-P. Berges, E.~Teng, A.~Cohen, J.~Harper, C.~Elion, C.~Goy,
  Y.~Gao, H.~Henry, M.~Mattar, and D.~Lange, ``Unity: A general platform for
  intelligent agents,'' 2020.

\bibitem{schulman2017proximal}
J.~Schulman, F.~Wolski, P.~Dhariwal, A.~Radford, and O.~Klimov, ``Proximal
  policy optimization algorithms,'' \emph{arXiv preprint arXiv:1707.06347},
  2017.

\bibitem{choset2000coverage}
H.~Choset, ``Coverage of known spaces: The boustrophedon cellular
  decomposition,'' \emph{Autonomous Robots}, vol.~9, no.~3, pp. 247--253, 2000.

\bibitem{tang2021mstc}
J.~Tang, C.~Sun, and X.~Zhang, ``{MSTC*}: Multi-robot coverage path planning
  under physical constraints,'' in \emph{2021 IEEE International Conference on
  Robotics and Automation}.\hskip 1em plus 0.5em minus 0.4em\relax IEEE, 2021.

\end{thebibliography}

\end{document}